\documentclass[final,5p,times,twocolumn]{elsarticle}

\usepackage{graphicx}
\usepackage{amsmath} 
\usepackage{framed,multirow}

\usepackage{amssymb}
\usepackage{latexsym}

\usepackage{url}
\usepackage{color}
\usepackage{amsmath}
\usepackage{amssymb}
\usepackage{amsfonts}
\usepackage{siunitx}
\usepackage{dcolumn}
\usepackage{bm}

\usepackage{hyperref}
\hypersetup{pdfauthor={Name}}

\usepackage{graphicx}
\usepackage{threeparttable}
\usepackage{booktabs}
\usepackage{multirow}
\usepackage{tabularx}

\begin{document}

\begin{frontmatter}
\title{Deep Bayesian Active Learning-to-Rank with Relative Annotation for Estimation of Ulcerative Colitis Severity}
\author[1]{Takeaki Kadota\corref{cor1}}
\cortext[cor1]{Corresponding author: Takeaki Kadota}
\ead{takeaki.kadota@human.ait.kyushu-u.ac.jp}
\author[2]{Hideaki Hayashi}
\author[1,3]{Ryoma Bise}
\author[4]{Kiyohito Tanaka}
\author[1,3]{Seiichi Uchida}

\address[1]{Department of Advanced Information Technology,
Kyushu University, 744, Motooka, Nishi-ku, Fukuoka-shi, Fukuoka, 819-0395, Japan}
\address[2]{Institute for Datability Science, Osaka University, 2-8, Yamadaoka, Suita-shi, Osaka, 565-0871, Japan}
\address[3]{Research Center for Medical Bigdata,
National Institute of Informatics, 2-1-2, Hitotsubashi, Chiyoda-ku, Tokyo, 101-8430, Japan}
\address[4]{Department of Gastroenterology, Kyoto Second Red Cross Hospital, 
355-5, Haruobicho Kamigyo-ku, Kyoto-shi, Kyoto, 602-8026, Japan}

\begin{abstract}
Automatic image-based severity estimation is an important task in computer-aided diagnosis. Severity estimation by deep learning requires a large amount of training data to achieve a high performance. In general, severity estimation uses training data annotated with discrete (i.e., quantized) severity labels. Annotating discrete labels is often difficult in images with ambiguous severity, and the annotation cost is high. In contrast, relative annotation, in which the severity between a pair of images is compared, can avoid quantizing severity and thus makes it easier. We can estimate relative disease severity using a learning-to-rank framework with relative annotations, but relative annotation has the problem of the enormous number of pairs that can be annotated. Therefore, the selection of appropriate pairs is essential for relative annotation. In this paper, we propose a deep Bayesian active learning-to-rank that automatically selects appropriate pairs for relative annotation. Our method preferentially annotates unlabeled pairs with high learning efficiency from the model uncertainty of the samples. We prove the theoretical basis for adapting Bayesian neural networks to pairwise learning-to-rank and demonstrate the efficiency of our method through experiments on endoscopic images of ulcerative colitis on both private and public datasets. We also show that our method achieves a high performance under conditions of significant class imbalance because it automatically selects samples from the minority classes.
\end{abstract}

\end{frontmatter}

%text of the article
%% main text
%--------------------------------------------------------
\section{Introduction\label{sec:intro}}
%--------------------------------------------------------
Automatic image-based severity estimation is important to assist medical doctors in clinical practice. Deep learning has been applied to many disease severity estimations~\citep{Cho2019, Klang2021, Takenaka2020}. Severity estimation using deep learning requires the collection of a large amount of training data annotated with severity labels by medical experts. Since medical experts need to carefully identify disease severity for many images, creating training data demands laborious efforts.\par
\begin{figure}[t]
\centering
\includegraphics[width=\linewidth]{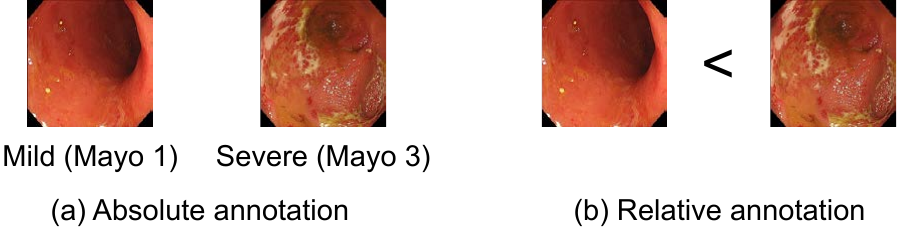}
\caption{Absolute and relative annotations.} 
\label{fig1}
\end{figure}

Standard annotations (hereafter referred to as \textit{absolute annotations}) represent disease severity as discretized severity labels. Figure~\ref{fig1}(a) shows the absolute annotations for endoscopic images of ulcerative colitis (UC). Absolute annotation can be difficult even for medical experts. This is because disease severity is inherently continuous, and when expressed in discrete severity levels, the levels can be ambiguous in intermediate cases. For example, when medical experts annotate medical images to be classified into four severity levels (0, 1, 2, and 3), they frequently encounter images near the intermediate severity levels (0.5, 1.5, and 2.5). Assigning discrete severity levels to these images is a time-consuming task that potentially leads to variability in decision outcomes. It has also been reported that absolute annotation has high variability not only between different medical experts but also within the same medical expert~\citep{Hirai2008}.\par
\textit{Relative annotation} is a promising alternative to absolute annotation in that it offers an easier process. Figure~\ref{fig1}(b) shows the relative annotations for UC endoscopic images, where we compare the severity of the two images and attach relative labels that indicate the result of the comparison. Relative annotation tends to be easier for annotators compared to absolute annotation, and it helps reduce subjective bias. Consequently, relative annotation leads to less variability in decision outcomes across different annotators. Relative annotation has been used for pairwise learning-to-rank (LTR) methods, mainly for ranking tasks in information retrieval~\citep{Carterette2006, Liu2009, Leaman2013, Hofmann2013}. In computer vision, relative annotation is also used in image analysis applications because it is easy to perform and is stable when labeling continuously changing data~\citep{Parikh2011}. Recently, relative annotation has been applied to medical images, and reports have indicated that it reduces annotation costs and labeling errors~\citep{Kadota_Access_2022, Saibro2022}.\par

\begin{figure*}[t]
\centering
\includegraphics[scale=0.63]{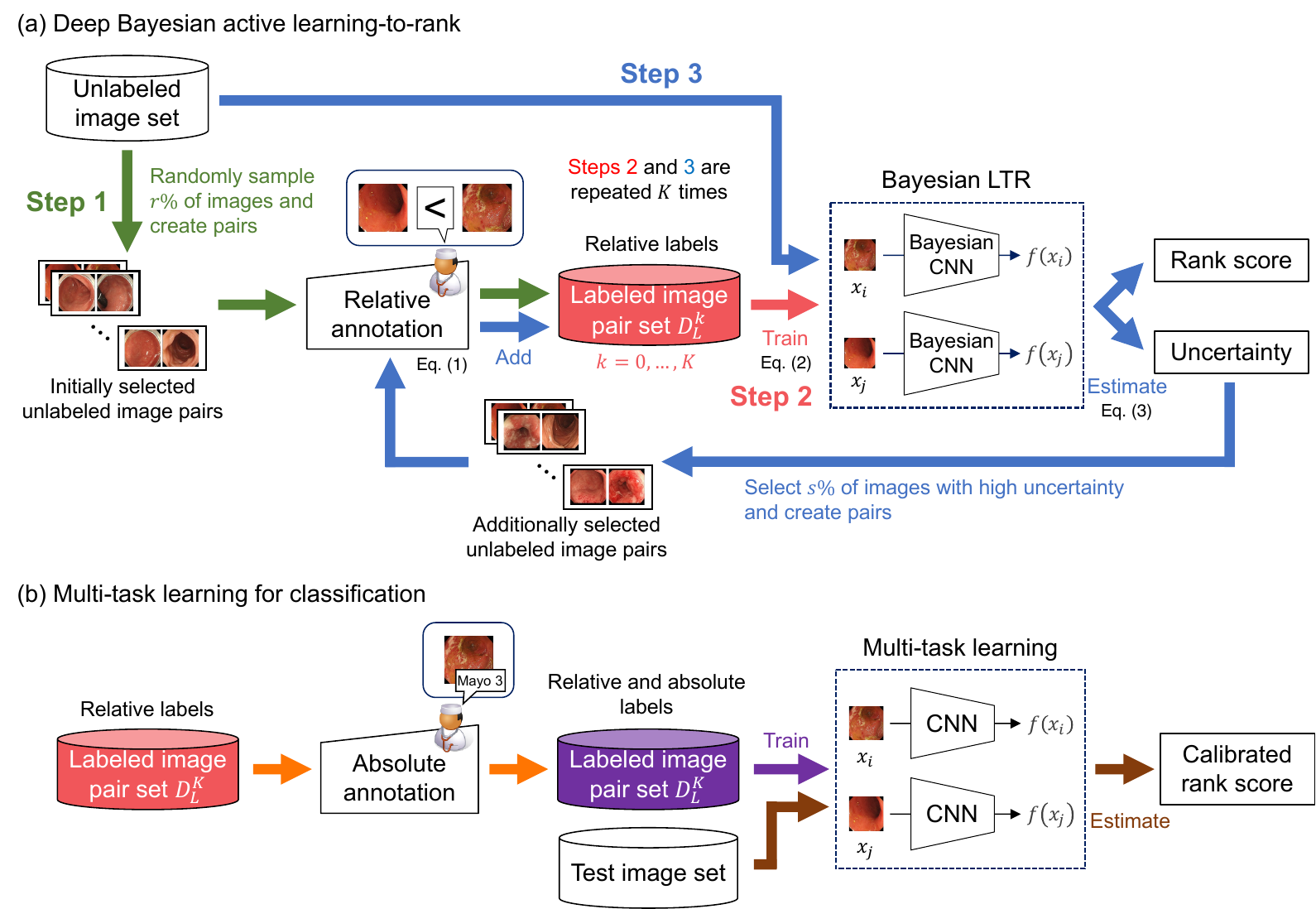}
\caption{(a) Deep Bayesian active learning-to-rank for relative severity estimation; step 1 (green arrows): generating a small number of pairs using randomly selected images from an unlabeled image set and annotating these pairs for the initial training; step 2 (red arrow): training the Bayesian CNN using the labeled image pair set; step 3 (blue arrows): selecting high-uncertainty images from the unlabeled image set to create pairs and attaching relative labels to the pairs. (b) Multi-task learning for severity classification.}
\label{fig2}
\end{figure*}

Image pair datasets with relative annotation can be used for ranking tasks to estimate the order of image severity. Given an image pair $(x_i, x_j)$, where $x_i$ has a higher severity than $x_j$, a ranking function $f(x)$, which outputs a scalar value called a rank score, is trained such that $f(x_i)>f(x_j)$ is satisfied. Since $f(x)$ gives a higher rank score to an image with higher severity, it can be used as the severity of $x$. Therefore, we can estimate the order of severity for multiple images by comparing their rank scores. Furthermore, a calibration method using a small number of absolute annotations allows $f(x)$ to be used in the classification task as the absolute severity instead of the relative severity~\citep{Kadota_Access_2022}. In addition, $f(x)$ can be used for two-class classification when medical experts determine the threshold for $f(x)$~\citep{Saibro2022}.\par
A critical challenge in LTR with relative annotation is the need to carefully select and annotate highly effective pairs from all possible pairs for learning. In relative annotation, there are $N(N-1)/2$ possible pairs for $N$ image samples. Even though individual relative annotation is easy, it is practically difficult to annotate all possible pairs. In addition, even if all pairs were used for training, the training time would be considerably longer, and less effective pairs for learning could reduce performance. Therefore, it is essential to preferentially select and annotate pairs with high learning effectiveness suitable for severity estimation from all pairs.\par
In this paper, we address the problem of pair selection for relative annotations by using an active learning framework based on a Bayesian convolutional neural network (Bayesian CNN). For a trained model, Bayesian CNN estimates the uncertainty of the sample. We employed a Bayesian CNN because it can estimate uncertainty by applying Monte Carlo (MC) dropout~\citep{Gal2016} without changing the network structure for the pairwise LTR framework. The proposed method features \textit{active learning}, in which the Bayesian CNN is introduced into the LTR framework to find pairs with high uncertainty and gradually add relative annotations to them. The experiments described in later sections show that the proposed method achieves high performance on training data with significantly fewer pairs than $N(N-1)/2$.\par
We also provide theoretical justification for the uncertainty estimation in LTR based on the Bayesian CNN. Our method applies MC dropout to CNNs with a Siamese network structure of pairwise ranking approaches. The application of MC dropout to CNNs in regression and classification tasks has already been demonstrated~\citep{Gal2017}. Here, we prove that MC dropout is equally effective for CNNs in a Siamese network structure in the ranking task. Our findings indicate that the proposed method may be effective not only for severity estimation with relative annotation but also for various other applications. Furthermore, through experiments using both private and public UC endoscopic image datasets, we demonstrate the effectiveness of the proposed method in severity estimation. Medical image datasets generally have class imbalance because there are more images with normal or mild disease than those with severe disease. We found that the proposed method preferentially selects important samples from the minority classes and thereby reduces the class imbalance.\par
The main contributions of this paper are as follows:
\begin{enumerate}
\item We propose an active learning method that introduces Bayesian CNN into a learning-to-rank framework to address the pair selection problem, an essential challenge in relative annotation.
\item We theoretically demonstrate the applicability of MC dropout in estimating uncertainty for a pairwise LTR task using a Bayesian Siamese neural network (NN).
\item We experimentally demonstrate that the proposed method improves the severity estimation performance by selecting pairs with high learning efficiency on a UC endoscopy image dataset. In addition, we verify the generalization ability of the proposed method using a public dataset. We also confirm that the proposed method is robust to class imbalances by selecting minority but important samples with high priority.
\item We prove the usefulness of the proposed method in a severity classification task that classifies each image into one of the discretized severity levels, which is a common task in medical image diagnosis.
\end{enumerate}
A preliminary version of this paper was published as a conference paper at MIUA2022~\citep{Kadota_MIUA_2022}. In the current version, we add the following new and significant contributions.
\begin{itemize}
\item We prove that MC dropout can be applied to pairwise LTR by deriving the variational lower bounds in pairwise LTR with Siamese network structures. We then show how this allows us to treat the Siamese network structure with dropout as a Bayesian model and obtain uncertainty estimates for its features (corresponds to Contribution~2).
\item We experimentally verify the effectiveness of our method in the task of relative severity estimation using a public dataset of UC endoscopic images (as a part of Contribution~3).
\item To use the proposed method in the multi-class classification tasks commonly used in clinical practice to make treatment decisions, we integrate a technique to estimate the discrete severity levels by calibrating the rank score to the UC severity score into the proposed method (corresponds to Contribution~4).
\end{itemize}

%--------------------------------------------------------
\section{Related work\label{sec:related}}
%--------------------------------------------------------
\subsection{Disease severity estimation\label{subsec:Related disease class}}
\subsubsection{Severity estimation based on endoscopic images}
Many methods have been proposed to estimate disease severity based on endoscopic images. Liu et al. introduced a classification method that estimates three endoscopic severity levels of esophageal cancer: normal, premalignant, and cancerous~\citep{Liu2020}. Klang et al. used an NN to estimate five severity levels of Crohn's disease using capsule endoscopy ~\citep{Klang2021}. However, these methods come with high annotation costs due to the need for absolute annotation that attaches discrete levels to continuously changing lesions on endoscopic images. Cho et al. used endoscopic images of pathologically confirmed gastric lesions to estimate five severity levels of gastric cancer~\citep{Cho2019}. Although they demonstrated the feasibility of estimating pathologically diagnosed severity based on endoscopic images, their annotation process involved obtaining pathological images through biopsies, leading to considerable annotation costs.\par
\subsubsection{Ulcerative colitis (UC) severity estimation}
Several studies have investigated deep learning applications to UC severity estimation. Takenaka et al. used a dataset including the Ulcerative Colitis Endoscopic Index of Severity (UCEIS) for severity estimation~\citep{Takenaka2020}. UCEIS is a discrete severity score ranging from zero to eight points based on vascular pattern, bleeding, and erosion/ulceration, but it incurs a very high annotation cost. Palot et al. proposed an ordinal regression-based method to estimate the Mayo score, which categorizes UC severity into four levels~\citep{Polat2022}. They proposed a unique loss function that focuses on severity order, but the annotation cost is high because they use discrete severity levels as ground truth. To avoid the costly annotation of all captured images, Schwab et al. proposed a weakly supervised learning method that estimates UC severity~\citep{Stidham2019}. Becker et al. proposed a method to reduce annotation costs by automatically extracting images suitable for UC severity scoring from endoscopic video frames~\citep{Becker2021}. Although these studies tackle the issue of annotation costs in medical image datasets, they do not inherently solve the quantization error in disease severity, which is the root cause of rising annotation costs.\par

\subsection{Learning to rank (LTR) with relative annotation\label{subsec:Related relative annotation}}
\subsubsection{LTR for general image analysis}
LTR with relative annotation has been applied to image analyses such as attribute evaluation and image quality assessment (IQA). Parikh et al. proposed a relative attribute model that predicts attribute similarity to images instead of a general classification model that predicts the category of an attribute~\citep{Parikh2011}. In addition, Souri et al. proposed using a CNN-based architecture to predict the strength of relative attributes~\citep{Souri2016}. Ma et al. used a CNN-based pairwise LTR architecture for IQA~\citep{Ma2017}, and Liu et al. used a Siamese network structure to predict the quality ranking of images using a dataset featuring relative annotations for IQA~\citep{Liu2017}. These studies focus on IQA but do not address the critical issue of relative annotation, which causes an enormous increase in the number of annotations due to pair creation.\par

\subsubsection{LTR for medical image analysis}
Recent studies have reported the successful application of LTR with relative annotation to the image analysis of continuously changing lesions. Kalpathy-Cramer et al. modeled relative severity using relative annotations and proposed continuous severity scores for the retinopathy of prematurity~\citep{Kalpathy2016}. They found poor absolute agreements on classification but good relative agreements on disease severity in expert diagnoses. Li et al. developed a Siamese NN approach to assess changes between disease severity at a single time point and longitudinal patient visits, focusing on continuous disease changes~\citep{Li2020}. Lyu et al. proposed automatically selecting high-quality images based on image quality ranking using pairwise LTR~\citep{Lyu2021}. As mentioned above, using LTR with relative annotation, Kadota et al. found that annotation costs could be reduced~\citep{Kadota_Access_2022}, while Saibro et al. found that labeling errors could be reduced~\citep{Saibro2022}. However, although these studies show the effectiveness of relative annotation, they do not solve the issue of pair selection in relative annotation.\par

\subsection{Active learning\label{subsec:Active learning}}
\subsubsection{Diversity-based sampling}
Diversity-based sampling, one of the main techniques in active learning, is a selection strategy to efficiently find samples representative of the entire data distribution in the feature space. Dasgupta et al. proposed efficient sample selection using hierarchical clustering~\citep{Dasgupta2008}. Sener et al.~reformulated active learning as core-set selection and examined the core-sets by solving a greedy k-center problem~\citep{Sener2018}. Thapa et al. applied an active learning method based on core-set to semantic segmentation and depth estimation of endoscopic images~\citep{thapa2022}. Sourati et al. proposed a sampling based on Fisher information for CNNs~\citep{Sourati2018}. Smailagic et al. proposed a selection method for unlabeled samples using a distance function in the feature space~\citep{Smailagic2020}. The sample selection implemented in these studies assumes absolute annotation and does not consider relative annotation that creates pairs.\par

\subsubsection{Uncertainty-based sampling}
Many techniques for uncertainty-based sampling have been proposed in medical image analysis to select samples with high uncertainty as informative samples. Yang et al. and Gorriz et al. used pixel-wise sample uncertainty to determine effective annotation regions in their segmentation tasks~\citep{Yang2017, Gorriz2017}. Tang et al. used active learning with uncertainty selection and pseudo labels for the classification and segmentation of endoscopic images~\citep{TANG2023106723}. Wen et al. proposed a sample selection method for pathology images using patch-wise uncertainty~\citep{Wen2018}. Nair et al. used voxel-wise uncertainty measures for 3D lesion segmentation~\citep{Nair2020}. These methods do not assume any LTRs with relative annotations because they deal with segmentation or detection tasks.\par

In the field of natural language processing, Wang et al. have recently reported a method for applying a Bayesian CNN to LTR~\citep{Wang2021}. Although their method seems similar to ours, their purpose, structure, and application are quite different. Specifically, their objective is to rank sentences (answers) for a given sentence (query) according to their relevance. For this purpose, their network always takes two inputs (e.g., $f(x_i, x_j)$), whereas our network takes one input (e.g., $f(x_i)$). These differences make it impossible to use their method for active relative annotation tasks and thus to compare our method with theirs.

%--------------------------------------------------------
\section{Deep Bayesian active learning-to-rank\label{sec:method}}
%--------------------------------------------------------
In relative annotation, labeled image pair sets $\mathcal{D}^k_\mathrm{L} = \{(\bm{x}_i, \bm{x}_j, C_{i,j})\}$, where $\bm{x}_i, \bm{x}_j$ are input images, $C_{i,j}$ is the ground truth of the pair (relative label), and $k$ is the number of repetitions, are obtained by repeatedly selecting images from the unlabeled image set $\mathcal{D}_\mathrm{U} = \{\bm{x}_i\}$ to create pairs that are then annotated by medical experts. Our purpose is to achieve a high performance with a small number of pairs by selecting images from the unlabeled image set that are highly effective in learning.\par

We employed a Bayesian CNN for uncertainty estimation in pairwise LTR to select highly effective samples for training. This is because MC dropout with the Bayesian CNN can estimate uncertainty without changing the network structure of the pairwise LTR, which uses relative annotation. While ensemble-based methods can also be used to estimate uncertainty in deep neural networks~\citep{Beluch2018}, they require substantial computational resources due to the need to train multiple networks, and the cost would be further amplified in Siamese network structures. For these reasons, we adopted the Bayesian CNN for uncertainty estimation in active learning for pairwise LTR.\par

As shown in Fig.~\ref{fig2}(a), the proposed method consists of Bayesian inference based on the LTR algorithm for disease severity estimation and uncertainty-based active learning. While the Bayesian CNN is trained on the basis of the LTR algorithm, medical experts gradually add relative labels to the selected image pairs based on the uncertainty the Bayesian CNN provides. The proposed method consists of three steps. In step~1, we generate a small number of pairs using randomly selected images from an unlabeled image set. Medical experts annotate these pairs for the initial training. In step~2, we train the Bayesian CNN using the labeled image pair set to estimate the rank score and uncertainty of the individual training samples. In step~3, high-uncertainty images are selected from the unlabeled image set based on the estimated uncertainty to create pairs, and the medical expert attaches relative labels to the pairs. We repeat steps~2 and 3 $K$ times to train the Bayesian CNN while gradually increasing the training data by adding labeled image pairs.\par

\subsection{Preparing pairs for initial training (step~1) \label{sec:Step1}}
Relative annotation is generally performed by randomly selecting the number of pairs that can be annotated since this number increases quadratically with the image data. However, random selection results in many images of the majority class and can thus lead to a dataset with low learning efficiency due to many similar images. Therefore, the proposed method randomly selects as small a number of images as possible for the dataset in the initial training and prepares a labeled image pair set $\mathcal{D}^0_\mathrm{L}$ as follows. First, given a set of $N$ unlabeled images, randomly sample $R$ images, namely, $r\%$ of all images (i.e., $R = rN/100$ image samples). Then, the set of $R$ pairs is formed by randomly selecting from $R-1$ samples for each $R$ sample. The strategy for pair formation is arbitrary. Here, we form $R$ pairs instead of all possible $R(R-1)/2$ pairs to limit the number of pairs to be annotated. This strategy is used to avoid annotating $O(R^2)$ and thus to annotate all samples of $R$ at least once. Medical experts attach relative labels to these image pairs. Relative labels $C_{i,j}$ are defined for image pairs $(\bm{x}_i,\bm{x}_j)$ as follows:
\begin{equation}
    C_{i,j}=
    \left\{ \begin{array}{ll}
    1, &  \mbox{if $\bm{x}_i$ has more severity than $\bm{x}_j$,} \\
    0.5, & \mbox{else if $\bm{x}_i$ and $\bm{x}_j$ have equal severity,}  \\
    0, & \mbox{otherwise.}  \\
    \end{array} \right.
\end{equation}
Through this step, we obtain the annotated image pair sets $\mathcal{D}^0_\mathrm{L} = \{(\bm{x}_i, \bm{x}_j, C_{i,j})\}$ and $|\mathcal{D}^0_\mathrm{L}|=R$ for initial training.

\subsection{Training a Bayesian CNN (step~2) \label{sec:Step2}}
In active learning, uncertainty-based sampling is generally used to reduce the annotation cost for training data. Uncertainty-based sampling preferentially selects highly effective samples for learning. Gal et al. proposed approximate Bayesian inference with MC dropout for applying this sampling method to deep learning~\citep{Gal2016, Gal2017}. Their sampling strategy is to obtain the model uncertainty of samples in regression and classification tasks through a Bayesian CNN. While they provide theoretical proof that MC dropout functions as a Bayesian CNN, their study did not consider it in Siamese network structures for ranking tasks. 

LTR with relative annotations generally uses a pairwise LTR algorithm with a Siamese network structure. We propose an uncertainty-based active learning method that applies Bayesian CNN to Siamese network structures for ranking tasks. A Bayesian CNN is trained with a labeled image pair set $\mathcal{D}^0_\mathrm{L}$ and obtained as a ranking function. The obtained CNN outputs a rank score that represents the input image's severity order and the uncertainty of the rank score. We used RankNet~\citep{Burges2005}, a pairwise ranking method with a Siamese network structure, for the LTR algorithm. We employed RankNet because it utilizes a neural network as the model, and uncertainty can be easily estimated by applying MC dropout. RankNet uses a probabilistic ranking cost function in training. The proposed method performs approximate Bayesian inference by applying MC dropout to the Siamese neural network.

Let $f( \cdot)$ be a ranking function that is a CNN with $L$-weighted layers. Given a single image $x$, the CNN returns a scalar value $f(x)$ as the ranking score for $x$. Let $\mathbf{W}_l$ be the $l$-th weight tensor of the CNN. The Bayesian CNN is trained on the mini-batch $\mathcal{M}$ sampled from $\mathcal{D}^0_\mathrm{L}$, with dropout performed using the loss function $\mathcal{L}_\mathcal{M}$ defined as follows:
\begin{flalign} 
    \mathcal{L}_\mathcal{M}^{\mathrm{rank}}&=-\sum_{(i, j)\in\mathcal{I}_\mathcal{M}}\left\{C_{i,j}\log{P}_{i,j}+(1-C_{i,j})\log(1-{P}_{i,j})\right\} \nonumber \\ 
    & \quad + \lambda \sum_{l=1}^L\|\mathbf{W}_l \|_F^2,
\label{eq:loss_function}
\end{flalign}
where $\mathcal{I}_\mathcal{M}$ is a set of index pairs by the elements in mini-batch $\mathcal{M}$, $P_{i,j} = \mathrm{sigmoid}(f(\bm{x}_i)-f(\bm{x}_j))$ is a probability obtained from output values, $\lambda$ is a constant value for weight decay, and $\|\cdot\|_F$ is a Frobenius norm. The first term in Eq.~(\ref{eq:loss_function}) is a probabilistic ranking loss function~\citep{Burges2005} for the CNN to learn rank scores. The loss function is a cross-entropy loss defined by the target probability $C_{i,j}$ as ground truth and the probability $P_{i,j}$ obtained from the output values. The second term in Eq.~(\ref{eq:loss_function}) is a weight regularization term that can be derived from the Kullback-Leibler divergence between the approximate posterior and the posterior of the CNN weights~\citep{Gal2016}. The CNN is trained to minimize the loss function $\mathcal{L}_\mathcal{M}$ for each mini-batch while performing dropout. With a probability of $p_\mathrm{dropout}$, if the binary random variable takes one, it is sampled for every unit in the CNN at each forward calculation, and if the corresponding binary variable takes zero, the output of the unit is set to zero. 

The rank score of the unlabeled image $\bm{x}^*$ is the average of the predictions by the trained Bayesian CNN calculated as $y^*=\frac{1}{T}\sum_{t=1}^Tf(\bm{x}^*;\bm{\omega}_t)$ where $T$ is the number of sampling operations by MC dropout, $\bm{\omega}_t$ is the $t$-th realization of the set of CNN weights obtained by MC dropout, and $f(\cdot;\bm{\omega}_t)$ is the output of $f(\cdot)$ given a set of weights $\bm{\omega}_t$.

The prediction uncertainty obtained from the Bayesian CNN is defined as the variance of the posterior distribution of $y^*$. This uncertainty is used to select images for annotation in active learning and plays an important role in obtaining a highly effective dataset for learning. The variance of the posterior distribution, $\mathrm{Var}_{q(y^*|\bm{x}^*)}[y^*]$, which represents the uncertainty, is approximately obtained using MC dropout as follows:
\begin{flalign}
\mathrm{Var}_{q(y^*|\bm{x}^*)}[y^*] 
&\!\!=\!\!\mathbb{E}_{q(y^*|\bm{x}^*)}[(y^*)^2] - \left(\mathbb{E}_{q(y^*|\bm{x}^*)}[y^*]\right)^2 \nonumber \\
&\!\!\approx\!\! \frac{1}{T}\!\! \sum^T_{t=1}\!\left(f(\bm{x}^*;\! \bm{\omega}_t)\right)^2 \!-\! \left(\frac{1}{T}\!\!\sum^{T}_{t=1}\!f(\bm{x}^*;\! \bm{\omega}_t)\right)^2 \!\!+\! \mathrm{const.},
\label{eq:var_uncertainty}
\end{flalign}
where $q(y^*|\bm{x}^*)$ is the posterior distribution estimated by the model. In the next step, the absolute value of the uncertainty is not required, and thus, the constant term can be ignored.

\subsection{Uncertainty-based sample selection (step~3) \label{sec:Step3}}
The estimated uncertainty and relative annotations provide a new set of annotated image pairs. We use a trained Bayesian CNN to estimate the rank scores and associated uncertainties for the unlabeled images and select the top $s\%$ of images with high uncertainty. As in step~1, image pairs are generated from the selected images, and the medical expert annotates the image pairs with relative labels. The image pairs with newly attached relative labels are added to the current set of annotated image pairs $\mathcal{D}^0_\mathrm{L}$. The updated $\mathcal{D}^1_\mathrm{L}$ is used to retrain the Bayesian CNN. We repeat steps~2 and 3 $K$ times to increase the size of the annotated set $\mathcal{D}^k_\mathrm{L}$ ($k=0, \ldots, K$).

\subsection{Additional absolute annotation for multi-class classification \label{sec:additional}}
As shown in Fig.~\ref{fig2}(b), to apply the proposed method to a multi-class classification task, the training set with relative labels is additionally annotated with absolute labels for multi-task learning. The training set obtained in step~3 has relative labels attached to the pairs but no absolute labels attached to the individual images. For a classification task, absolute labels need to be used as ground truth, so we perform absolute annotation on each image to the obtained training set. In addition, to calibrate the rank score to the disease severity score, LTR and regression are trained simultaneously with absolute and relative labels by multi-task learning. The regression loss function is defined by the squared error loss function for each sample of the pairs as follows:
\begin{equation}
    L_\mathcal{M}^{\mathrm{reg}}=\sum_{(i, j)\in\mathcal{I}_\mathcal{M}}\left\{(f(x_{i})-A_{i})^{2}+(f(x_{j})-A_{j})^{2}\right\},
\end{equation}
where $A_i$ and $A_j$ are absolute labels of $\bm{x}_i$ and $\bm{x}_j$, respectively. The loss function of multi-task learning is defined as the sum of the LTR loss function $L_\mathcal{M}^{\mathrm{rank}}$ in Eq.~(\ref{eq:loss_function}) and the regression loss function $L_\mathcal{M}^{\mathrm{reg}}$.

\section{Theoretical analysis of Bayesian learning-to-rank\label{sec:proof}}
\subsection{Evaluating log evidence lower bound for ranking \label{subsec:mc dropout for ranking}}
In this section, we demonstrate that model uncertainty can be estimated by utilizing MC dropout in the context of LTR employing a Siamese network. The validity of the MC dropout-based uncertainty estimation for an NN was originally shown by Gal and Ghahramani on general classification and regression tasks~\citep{gal2016dropout}. Their approach involved obtaining an approximate distribution of the posterior distribution over the weights of the NN through variational inference, which is attributed to maximizing the log-evidence lower bound. This was then shown to be equivalent to minimizing the loss function (cross-entropy for classification or mean squared error for regression) with $L_2$ regularization and MC dropout. Our proof process follows that of the aforementioned general classification and regression cases, and we primarily establish the equivalency between a log-evidence lower bound and the loss function with $L_2$ regularization and MC dropout in the context of LTR with a Siamese network.

We consider the adaptation of MC dropout to the Siamese network structure for the case of an NN with a single hidden layer. We use the Siamese network structure consisting of single hidden layer NNs to simplify the explanations in this section, but the generalization to multi-layer NNs is straightforward. Let $\mathbf{W}_1$ and $\mathbf{W}_2$ be the weight matrices connecting the first layer to the hidden layer and the hidden layer to the output layer, respectively, and let $\bm{b}$ be the bias. Let $\mathbf{X}_1$ and $\mathbf{X}_2$ be training data matrices with the $n$-th training sample $(\bm{x}_1^n)^\top$ and $(\bm{x}_2^n)^\top$ ($n = 1, \ldots, N$) in the $n$-th row, respectively. The output data matrices for the inputs $\mathbf{X}_1$ and $\mathbf{X}_2$ are denoted by $\mathbf{Y}_1$ and $\mathbf{Y}_2$ that contain the $n$-th output vector $(\bm{y}_1^n)^\top$ and $(\bm{y}_2^n)^\top$ in the $n$-th row, respectively. Given data matrices $\mathbf{X}_1$, $\mathbf{X}_2$, $\mathbf{Y}_1$, and $\mathbf{Y}_2$, we estimate the ranking function $y=f(\bm{x})$ obtained by the pairwise LTR of the Siamese network structures. Let $\bm{c}=\left[c^{1}, \ldots, c^{N}\right]^\top$ be the relative label matrix with the $n$-th relative label $c^n$ for $\bm{x}_1^n$ and $\bm{x}_2^n$. Then, we can write the generative model for the ranking task as follows:
\begin{flalign}
    p(c \mid \mathbf{X}_1,\mathbf{X}_2)&\!=\!\!\int\!\! p(c \mid \mathbf{Y}_1,\mathbf{Y}_2) p(\mathbf{Y}_1,\mathbf{Y}_2 \mid \mathbf{X}_1,\mathbf{X}_2) \mathrm{d}\mathbf{Y}_1 \mathrm{d}\mathbf{Y}_2 \nonumber \\
    &\!=\!\!\int\!\! p(c \mid \mathbf{Y}_1,\mathbf{Y}_2) p(\mathbf{Y}_1,\mathbf{Y}_2 \mid \mathbf{X}_1,\mathbf{X}_2,\mathbf{W}_1, \mathbf{W}_2, \bm{b}) \nonumber \\ 
    &\!\quad\! \cdot p(\mathbf{W}_1, \mathbf{W}_2, \bm{b}) \mathrm{d}\mathbf{W}_1 \mathrm{d}\mathbf{W}_2 \mathrm{d}\bm{b} \mathrm{d}\mathbf{Y}_1 \mathrm{d}\mathbf{Y}_2, 
\label{eq:generative model}
\end{flalign}
where $W _1$ is a $Q \times U$ matrix derived from $Q$-dimensional inputs and $U$ hidden units, $\mathbf{W}_2$ is a $U \times D$ matrix derived from $U$ hidden units and $D$-dimensional outputs, and $\bm{b}$ is a $U$-dimensional vector of bias terms. 

From Eq.~(\ref{eq:generative model}), the log-evidence lower bound of the pairwise LTR can be written as follows (the calculation process is described in the Supplementary Materials):
\begin{flalign}
    \mathcal{L}_\mathrm{GP-VI} \coloneqq &\int p(\mathbf{Y}_1,\mathbf{Y}_2 \mid \mathbf{X}_1,\mathbf{X}_2,\mathbf{W}_1, \mathbf{W}_2, \bm{b}) q(\mathbf{W}_1, \mathbf{W}_2, \bm{b})\nonumber \\ 
    &\quad \cdot \log (p(\bm{c} \mid \mathbf{Y}_1,\mathbf{Y}_2)) \mathrm{d}\mathbf{W}_1 \mathrm{d}\mathbf{W}_2 \mathrm{d}\bm{b} \mathrm{d}\mathbf{Y}_1 \mathrm{d}\mathbf{Y}_2 \nonumber \\
    &\quad - \mathrm{KL}(q(\mathbf{W}_1, \mathbf{W}_2, \bm{b})||p(\mathbf{W}_1, \mathbf{W}_2, \bm{b})),
\label{eq:log-variable}
\end{flalign}
where $q(\mathbf{W}_1, \mathbf{W}_2, \bm{b})$ is the approximating variational distribution, and KL is the Kullback-Leibler divergence. 

The integrand of the first term in Eq.~(\ref{eq:log-variable}) can be rewritten as a sum:
\begin{equation}
    \begin{split}
    \log \left(p(\bm{c} \mid \mathbf{Y}_1,\mathbf{Y}_2)\right)=\sum_{n=1}^N \log (p(c^n \mid \bm{y}_1^n,\bm{y}_2^n)).
    \end{split}
\label{eq:sum}
\end{equation}

As a result, Eq.~(\ref{eq:log-variable}) is expressed as follows:
\begin{flalign}   
    \mathcal{L}_\mathrm{GP-VI} \coloneqq &\sum_{n=1}^N \int p(y_1^n,y_2^n \mid \bm{x}_1^n,\bm{x}_2^n,\mathbf{W}_1, \mathbf{W}_2, \bm{b}) q(\mathbf{W}_1, \mathbf{W}_2, \bm{b}) \nonumber\\ 
    &\quad \cdot \log (p(c^n \mid y_1^n,y_2^n)) \mathrm{d}\mathbf{W}_1 \mathrm{d}\mathbf{W}_2 \mathrm{d}\bm{b} \mathrm{d}y_1^n \mathrm{d}y_2^n \nonumber \\
    &\quad - \mathrm{KL}(q(\mathbf{W}_1, \mathbf{W}_2, \bm{b})||p(\mathbf{W}_1, \mathbf{W}_2, \bm{b})).
\label{eq:log-variable-sum}
\end{flalign}

The integrands in the sum of Eq.~(\ref{eq:log-variable-sum}) can be re-parameterized not to depend directly on $\mathbf{W}_1$, $\mathbf{W}_2$, and $\bm{b}$ but rather on the standard normal and Bernoulli distributions. Let $\bm{z}_1$ and $\bm{z}_2$ be binary vectors whose element follows the Bernoulli distribution, as $q(z_{1,q}) = \mathrm{Bernoulli}(p_1)$ with $p_1\in[0,1]$ for $q = 1, \ldots, Q$ and $q(z_{2,u}) = \mathrm{Bernoulli}(p_2)$ with $p_2\in[0,1]$ for $u = 1, \ldots, U$. 
Let $\bm{\epsilon}_1 \in \mathbb{R}^{Q\times U}$, $\bm{\epsilon}_2 \in \mathbb{R}^{U \times D}$, and $\bm{\epsilon} \in \mathbb{R}^{U}$ be random matrices and a vector whose element independently follows the standard normal distribution. We re-parameterize the integrands as follows:
\begin{equation}
    \begin{split}
    \mathbf{W}_1&=\mathbf{M}_1 \mathrm{diag}(\bm{z}_1) + \sigma \bm{\epsilon}_1,\\
    \mathbf{W}_2&=\mathbf{M}_2 \mathrm{diag}(\bm{z}_2) + \sigma \bm{\epsilon}_2,\\
    \bm{b}&=m + \sigma \bm{\epsilon},\\
    y^n_1&=\sqrt{\frac{1}{U}}\mathbf{W}_2^\top\phi(\mathbf{W}_1^\top \bm{x}^n_1 +\bm{b}),\\
    y^n_2&=\sqrt{\frac{1}{U}}\mathbf{W}_2^\top\phi(\mathbf{W}_1^\top\bm{x}^n_2 + \bm{b}),
    \end{split}
\label{eq:re-parameter}
\end{equation}
where $\mathbf{M}_1=[m_{q}]_{q=1}^{Q}$, $\mathbf{M}_2=[m_{u}]_{u=1}^{U}$, and $m$ are variational parameters, $\mathrm{diag}(\bm{z})$ is an operation that returns a diagonal matrix with the elements of vector $\bm{z}$ on the main diagonal, $\sigma > 0$ is a scalar, and $\phi(\cdot)$ is an element-wise nonlinear function.

Next, using Monte Carlo integration with a distinct single sample, we estimate each integral for all pair samples as follows:
\begin{flalign}
    \mathcal{L}_\mathrm{GP-MC} \!\!\coloneqq\!\! &\sum_{n=1}^N \!\log (p(c^n \!\mid\! \hat{y}_1^n(\bm{x}_1^n,\!\hat{\mathbf{W}}_1^n,\!\hat{\mathbf{W}}_2^n,\!\hat{\bm{b}}^n),\!\hat{y}_2^n(\bm{x}_2^n,\!\hat{\mathbf{W}}_1^n,\!\hat{\mathbf{W}}_2^n,\!\hat{\bm{b}}^n)))\nonumber \\
    &\quad - \mathrm{KL}(q(\mathbf{W}_1, \mathbf{W}_2, \bm{b})||p(\mathbf{W}_1, \mathbf{W}_2, \bm{b})),
\label{eq:Monte Carlo}
\end{flalign}
where $\hat{y}_1^n$, $\hat{y}_2^n$, $\hat{\mathbf{W}}_1^n$, $\hat{\mathbf{W}}_2^n$, and $\hat{\bm{b}}^n$ are the realizations of $y_1^n,y_2^n$, $\mathbf{W}_1^n$, $\mathbf{W}_2^n$, and $\bm{b}^n$ sampled on the basis of Eq.~(\ref{eq:re-parameter}). 

The first term of the sum in Eq.~(\ref{eq:Monte Carlo}) can be written from the prediction probabilities of RankNet as follows:
\begin{flalign}
    \log (p(c^n \mid \hat{y}_1^n,\hat{y}_2^n))
    &=\log(\mathrm{sigmoid}(\hat{y}_1^n-\hat{y}_2^n)) \nonumber \\
    &=\log \Bigg(\frac{1}{1+\exp(\hat{y}_2^n-\hat{y}_1^n)}\Bigg).
\label{eq:sum_ranknet}
\end{flalign}

Using Monte Carlo integration, we can approximate the KL divergence term with the variational parameters $\mathbf{M}_1$, $\mathbf{M}_2$, and $\bm{m}$ and the probabilities $p_1$, $p_2$ (the details of the approximation are described in the Supplementary Materials). Furthermore, we can scale the objective by $1/N$ and optimize it to yield the maximization objective as follows:
\begin{flalign}
    \mathcal{L}_\mathrm{GP-MC} &\propto \frac{1}{N} \sum_{n=1}^N \log (p(c^n \mid \hat{y}_1^n,\hat{y}_2^n)) \nonumber \\
    &\quad - \frac{p_1}{2N}||\mathbf{M}_1||_2^2 - \frac{p_2}{2N}||\mathbf{M}_2||_2^2 - \frac{1}{2N}||\bm{m}||_2^2.
\label{eq:max objective}
\end{flalign}

In the training of an NN, a regularization term is often added to the loss function. The $L_2$ regularization weighted by some weight decays $\lambda$ is often used, and we can obtain the following equation to minimize the objective:
\begin{equation}
    \mathcal{L}_\mathrm{dropout} \coloneqq E + \lambda_1||\mathbf{W}_1||_2^2 + \lambda_2||\mathbf{W}_2||_2^2 + \lambda_3||\bm{b}||_2^2,
\label{eq:dropout NN}
\end{equation}
where $E$ is the loss function.

The optimal parameters for maximizing Eq.~(\ref{eq:max objective}) lead to the same as those for minimizing Eq.~(\ref{eq:dropout NN}) if the weight decays in Eq.~(\ref{eq:dropout NN}) are properly determined. The two equations converge to the same limit with the correct stochastic optimizer. The above results prove that MC dropout can be applied to pairwise LTR with the Siamese network structure.

\subsection{Acquisition function for pairwise ranking \label{subsec:acquisition function}}

The predictive distribution of the Siamese network model is represented as a joint probability, such as $p(\mathbf{Y}_1, \mathbf{Y}_2 \mid \mathbf{X}_1, \mathbf{X}_2)$ in Eq.~(\ref{eq:generative model}). In addition, $\mathbf{Y}_1$ and $\mathbf{Y}_2$ are independent of each other, and thus the distribution can be expressed as follows:
\begin{flalign}
    &p(\mathbf{Y}_1,\mathbf{Y}_2 \mid \mathbf{X}_1,\mathbf{X}_2,\mathbf{W}_1, \mathbf{W}_2, \bm{b}) \nonumber \\ 
    &= p(\mathbf{Y}_1 \mid \mathbf{X}_1,\mathbf{W}_1, \mathbf{W}_2, b) p(\mathbf{Y}_2 \mid \mathbf{X}_2,\mathbf{W}_1, \mathbf{W}_2, \bm{b}).
\label{eq:Separate type}
\end{flalign}

Therefore, we define the variance of the posterior distribution of the output value, $y^*$, for the single input $x^*$, as shown in Eq.~(\ref{eq:var_uncertainty}), as an acquisition function of the model uncertainty.

%--------------------------------------------------------
\section{Experiment}
%--------------------------------------------------------
\newcolumntype{C}{>{\centering\arraybackslash}X}
\newcolumntype{R}{>{\raggedright\arraybackslash}X}
\newcolumntype{L}{>{\raggedleft\arraybackslash}X}
\begin{table*}[t]
  \centering
  \caption{Quantitative performance evaluation of the accuracy of estimating relative labels. The labeling ratio shows the percentage of relative labels used in training. The labeling ratio of $100\%$ indicates that the labels were created from all training data using the pairing method described in Section~\ref{sec:evaluation metrics}. `*' indicates a statistically significant difference between the proposed method and each compared method at $p<0.05$ by multiple statistical comparisons using McNemar's test.}
  \label{tb:accracy}
    \scalebox{1}{
    \begin{tabularx}{\hsize}{llCCCCCC}
     \hline
     \multirow{2}{*}{Data}&\multirow{2}{*}{Method}&\multirow{2}{*}{\begin{tabular}{c}Labeling\\ratio\end{tabular}}&\multirow{2}{*}{Overall}&\multicolumn{4}{c}{Neighboring}\\
     \cline{5-8}
     &&&&0--1&1--2&2--3&Mean\\
     \hline \hline
     \multirow{4}{*}{Private}&Baseline&50\%&0.861$^*$&0.827$^*$&0.837$^*$&0.628$^*$&0.763$^*$\\
     &Baseline (all data)&100\%&0.875&{\bf0.855}$^*$&0.870&0.635$^*$&0.785\\
     &Core-set&50\%&0.851$^*$&0.785&0.827$^*$&0.632$^*$&0.747$^*$\\
     &Proposed w/o UBS&50\%&0.856$^*$&0.818$^*$&0.842$^*$&0.634$^*$&0.763$^*$\\
     &Proposed&50\%&{\bf0.880}&0.787&{\bf0.871}&{\bf0.736}&{\bf0.797}\\
     \hline
     \multirow{4}{*}{Public}&Baseline&50\%&0.838$^*$&0.803&0.748$^*$&0.731$^*$&0.760$^*$\\
     &Baseline (all data)&100\%&0.878&{\bf0.827}$^*$&0.806&0.778$^*$&{\bf0.804}\\
     &Core-set&50\%&0.836$^*$&0.802&0.741$^*$&0.705$^*$&0.749$^*$\\
     &Proposed w/o UBS&50\%&0.857$^*$&0.804&0.776$^*$&0.749$^*$&0.776$^*$\\
     &Proposed&50\%&{\bf0.882}&0.793&{\bf0.813}&{\bf0.806}&{\bf0.804}\\
     \hline 
    \end{tabularx}
    }
\end{table*}

To evaluate the effectiveness of our method, we conducted experiments on the disease severity estimation task using one private and one public dataset for UC severity. In these experiments, we evaluated the estimation performance of the proposed method for two use cases. In the first use case, we examine the accuracy of estimating the relative labels, that is, correctly identifying the image with higher severity in a given endoscopic image pair. The objective of this evaluation is to determine the effectiveness of treatment and to assess changes in severity during follow-up in clinical practice. We quantitatively compare the proposed method to baseline methods and also analyze the relationship between uncertainty and class prior distribution for the datasets obtained by active learning.

In the second use case, we evaluate the performance of estimating UC severity levels in classification tasks. By attaching additional absolute labels to the dataset obtained by active learning, we estimate UC severity classification by multi-task learning~\citep{Kadota_Access_2022} combining LTR and regression. The performance and annotation cost of the proposed method are compared with those of conventional multi-class classification methods.

\vskip\baselineskip
\begin{figure}[t]
\centering
\includegraphics[scale=0.4]{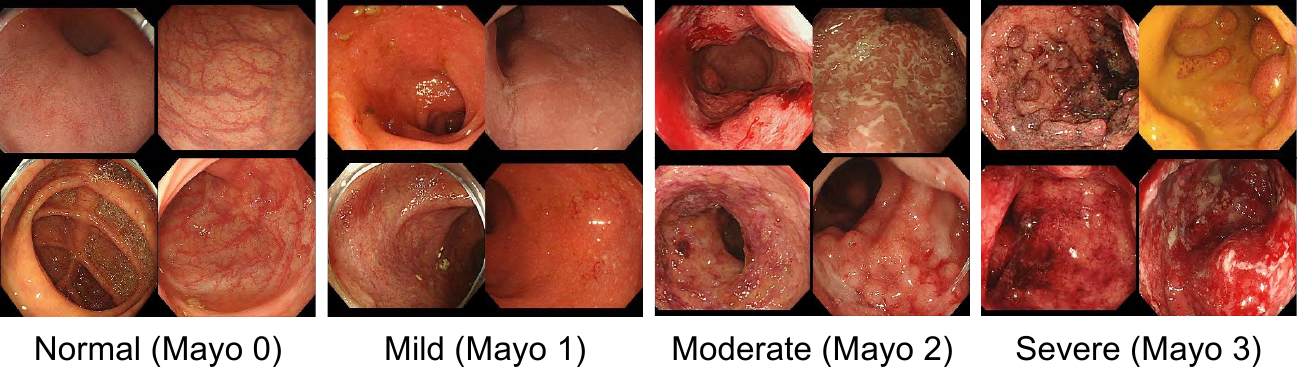}
\caption{Examples of endoscopic images of ulcerative colitis at each Mayo (severity).} \label{fig3}
\end{figure}
\subsection{Relative severity estimation}\label{sec:eval_rel}
\subsubsection{Dataset}\label{sec:dataset}
To examine the performance of the proposed method for relative label estimation, we used two datasets (one private and one public) with absolute severity labels for UC. 

The private dataset contains 10,265 endoscopic images from 388 ulcerative colitis (UC) patients at Kyoto Second Red Cross Hospital. The Ethical Review Committee of Kyoto Second Red Cross Hospital approved the experiments using the private dataset. Multiple medical experts carefully annotated a Mayo score, which determines UC severity on a four-point scale (Mayo 0--3) for each image. Figure~\ref{fig3} shows examples of endoscopic images in which the Mayo score was determined. Schroeder et al.~\citep{Schroeder1987} defined a scoring system to assess UC activity as Mayo scores in which the endoscopic findings of UC in the system are divided into four stages: Mayo~0 is a normal or inactive disease, Mayo~1 is a mild disease with erythema, decreased vascular pattern, and mild friability, Mayo~2 is a moderate disease with marked erythema, absent vascular pattern, collapse, and erosion, and Mayo~3 is a severe disease with spontaneous bleeding and ulceration. The private dataset has a large class imbalance and contains 6,678, 1,995, 1,395, and 197 samples for Mayo~0, 1, 2, and~3, respectively. Note that medical imaging datasets usually have class imbalances because the number of patients with normal or mild disease is typically larger than the number of patients with severe disease.

The public dataset we used is the LIMUC dataset~\citep{polat2022labeled} annotated with Mayo scores for UC. This dataset contains 11,276 images from 564 patients, all annotated by at least two medical experts. The public dataset also has class imbalance: Mayo~0, Mayo~1, Mayo~2, and Mayo~3 are 6,105, 3,052, 1,254, and 865, respectively. In principle, the same settings were used in the evaluation of all experiments with the private dataset.

\subsubsection{Evaluation metrics}\label{sec:evaluation metrics}
We defined the accuracy of relative label estimation, namely, the percentage of relative labels correctly estimated for pairs, as a performance metric. In this experiment, we created pairs of two images and attached a relative label to each pair based on the Mayo score. Estimated relative labels were attached by comparing the rank scores of the image pairs. According to Li et al., the severity rank score obtained from the Siamese network can provide a continuous measure of the change in disease severity between images~\citep{Li2020}.

The number of pairs ($O(N^2)$) that can be created from the $N$ samples is too large when creating image pairs, so we limited the number of pairs for random sampling as follows. By selecting one sample from $N-1$ samples for each $N$ sample, we created $N$ pairs (instead of $O(N^2)$ pairs). We also utilized this pair creation method in Section~\ref{sec:method} for selecting the initial set of $R$ samples. This setting, which is typical for pairwise LTR evaluations~\citep{Kadota_Access_2022,Xu2021,2019You}, was used for all pair creations.

We performed five-fold cross-validation on all methods. The datasets were divided into training ($60\%$), validation ($20\%$), and test ($20\%$) sets using patient-based sampling to ensure that images from the same patient were not included in different sets. For a fair evaluation, we created the image pairs within each set after dividing the training, validation, and test sets. Therefore, there were no duplicate images or image pairs between each set.
\par

\subsubsection{Implementation details}
The experimental environment was Ubuntu 18.04 and two NVIDIA TITAN RTX 24GB GPUs. We implemented our method with Tensorflow 1.13.1 and Keras 2.2.4. We used the DenseNet-169 structure ~\citep{Huang2017} as the backbone of the Bayesian CNN. We trained the CNN with dropout ($p_\mathrm{dropout}$ = 0.2) and weight decay ($\lambda$ = 1$\times$10$^{-4}$) settings in the convolutional and fully connected layers. We used the Adam optimizer to optimize weight parameters. The initial learning rates were set to 1$\times$10$^{-5}$ for the private dataset and 2$\times$10$^{-5}$ for the public dataset. All images in the datasets were resized to $224 \times 224$ pixels and normalized to values between 0 and 255.

The hyperparameters for active learning in all experiments were set to $K=6$ for the number of iterations, $s=5\%$ for the selection rate from the training data at each sampling, $r=20\%$ for the selection rate from the training data when sampling the initial training data, and $T=30$ for the number of estimates for uncertainty estimation. The initial annotation ratio $r=20\%$ was selected as a realistically low-cost option in the relative annotation scenario. The annotation ratio of the labeled dataset after all iterations ($6$ in this case) was $50\%$ ($r+sK=20+ 5\times 6$), assuming $100\%$ for the case with all training data. 

\subsubsection{Compared methods}\label{sec:comparison}
The proposed method was compared with four methods, which are referred to as `baseline,' `baseline (all data),' `proposed w/o UBS,' and `Core-set'~\citep{Sener2018}. The baseline was trained by randomly sampling $r+sK$ pairs of training data (the same number as for the proposed method). The baseline (all data) was trained using $N$ pairs created using all training samples. The annotation ratio of the baseline (all data) was $100\%$, which is twice the maximum annotation ratio of the proposed method (that is, $r+sK=50\%$). The proposed w/o UBS was trained with increasing training data by $K$ iterations using random sampling, instead of uncertainty-based sampling (UBS) to confirm the effect of UBS. Core-set represents a diversity-based sampling strategy that was used in place of UBS to assess its efficacy. Core-set utilized 1664-dimensional features extracted from DenseNet-169 trained on ImageNet. Note that all methods shared the same backbone architecture, DenseNet-169-based CNN, to ensure a fair comparison. For all methods, rank scores were calculated based on the average of the predictions estimated $T=30$ times using Bayesian CNN.

\subsubsection{Quantitative evaluation with test data}\label{sec:eval}
\begin{figure}[t]
\centering
\includegraphics[scale=0.6]{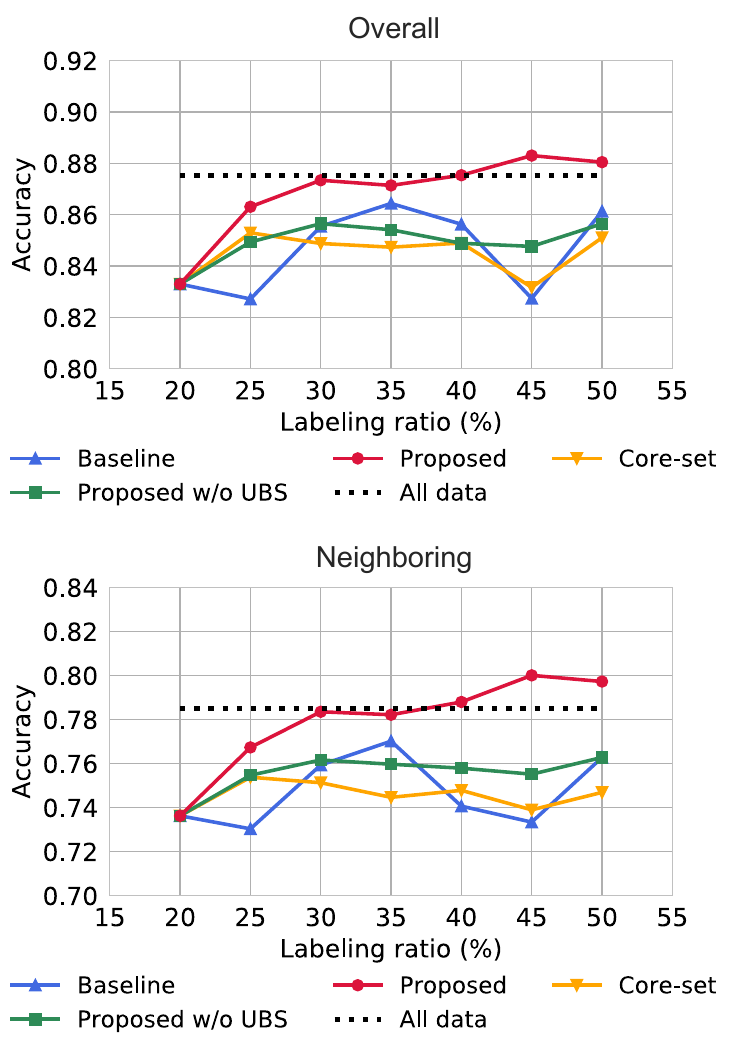}
\caption{Accuracy of relative label estimates for baseline (blue), Core-set (orange), proposed w/o UBS (green), and proposed method (red) at each labeling ratio. The black dotted line indicates the result of baseline (all data).}
\label{fig4}
\end{figure}

\begin{figure*}[t]
\centering
\includegraphics[scale=0.65]{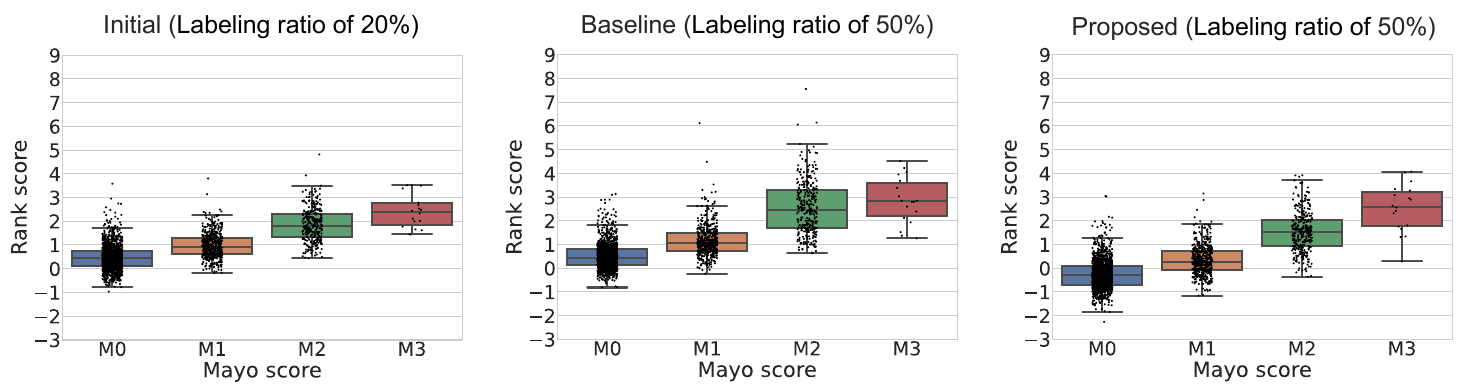}
\caption{Box plots of estimated rank scores at each Mayo score. The initial labeling ratio was measured with $20\%$ (iteration $K=0$). The results of the baseline and the proposed method estimates were measured with a labeling ratio of $50\%$ (iteration $K=6$). The estimate is considered reasonable if there is little overlap in the distribution of rank scores for each Mayo score.}
\label{fig5}
\end{figure*}

\begin{figure*}[t]
\centering
\includegraphics[scale=0.63]{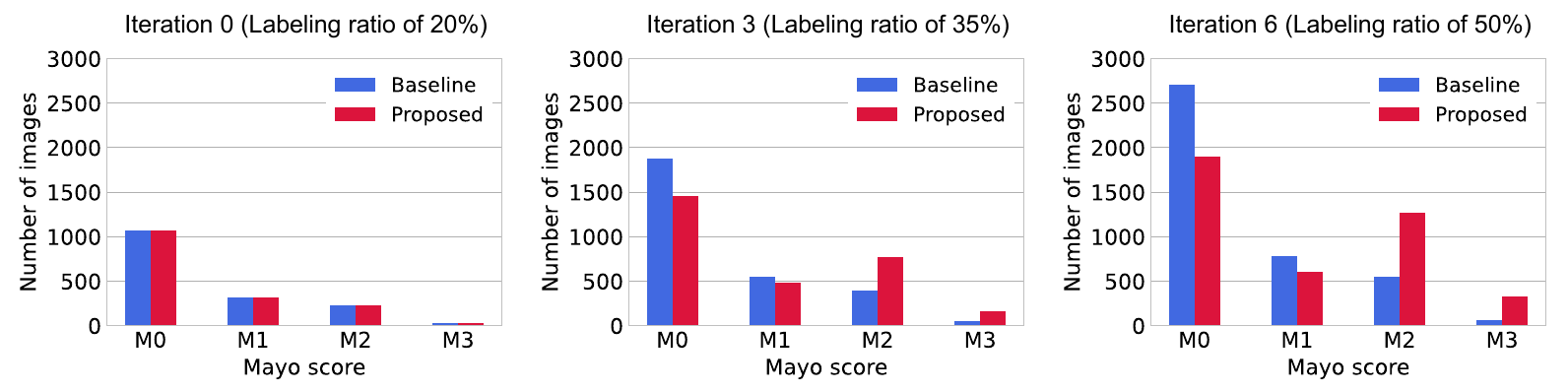}
\caption{Class proportions of each Mayo score for accumulated sampled images at iterations $K=0$ (labeling ratio of $20\%)$, $K=3$ (labeling ratio of $35\%$), and $K=6$ (labeling ratio of $50\%$). For a labeling ratio of $20\%$, the class proportions were the same for the baseline and the proposed method. The proposed method mitigates the class imbalance problem by selecting more samples from minority classes (Mayo 2 and~3).}
\label{fig6}
\end{figure*}

\begin{figure}[t]
\centering
\includegraphics[scale=0.75]{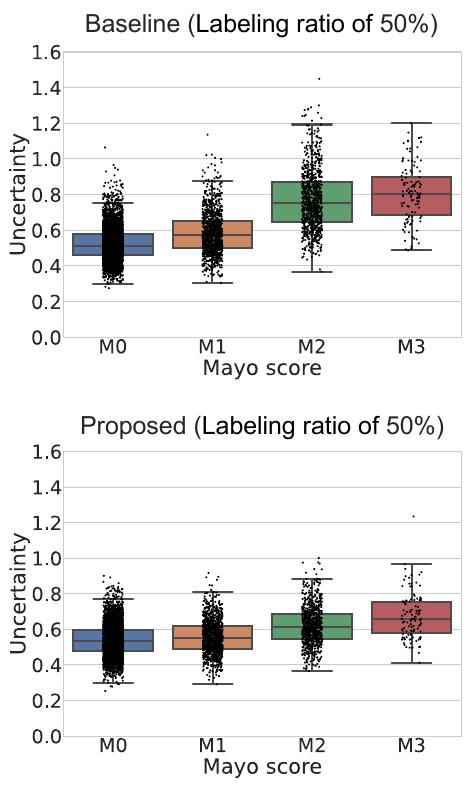}
\caption{Box plots of model uncertainty for each Mayo score for the baseline and the proposed method. The performance of each method was measured at a labeling rate of $50\%$. In the baseline, the minority classes (Mayo~2 and~3) had higher uncertainty than the majority class. The proposed method mitigated the class imbalance and reduced the uncertainty in Mayo~2 and~3.}
\label{fig7}
\end{figure}
We evaluated the accuracy of the relative severity estimation between the image pairs by comparing the estimated rank scores. Specifically, given a pair $(x_i, x_j)$ where $x_i$ is more severe than $x_j$, the estimate is considered ``accurate'' if $f(x_i)>f(x_j)$. 

Relative severity estimation is typically utilized in clinical practice to assess changes in disease severity. In general, it is difficult to compare severity when the difference in severity between pairs is small. Therefore, we prepared two types of test sets according to the difference in severity between pairs. Note that these test sets are obtained from virtual relative annotations with relative labels attached using Mayo labels.

\textbf{``Overall'' case:} We prepared a randomly paired test set from all Mayo scores. Specifically, we selected images so that the number of samples for each Mayo score was equal and randomly created pairs from the selected images. This test set contains many easy pairs, such as pairs with Mayo~0 (normal images) and Mayo~3 (severe images), which have very different degrees of severity, as shown in Fig.~\ref{fig3}. This test set was used to evaluate the overall performance of each method.

\textbf{``Neighboring'' case:} We created a neighboring pair test set with Mayo~0--1, Mayo~1--2, and Mayo~2--3, paired from neighboring Mayo score images. Neighboring pairs are similar in severity, making it difficult to determine their relative severity. In clinical applications, it is important to achieve high accuracy in cases where severity comparison is difficult. This test set evaluates the performance of each method in difficult cases.

Table~\ref{tb:accracy} shows the mean accuracy of relative label estimation by each method in five-fold cross-validation. The labeling ratio column shows the percentage of relative labels used in training. The labeling ratio of $100\%$ indicates that the labels were created from all training data using the pairing method described in Section~\ref{sec:evaluation metrics}. `*' denotes a statistically significant difference at $p<0.05$ by McNemar's test using Holm's method of multiple statistical comparisons.

In the results for the ``Overall'' test set, the proposed method achieved a higher performance than all other methods. In particular, it outperformed the baseline (all data) despite using half the amount of training data. The performances of the baseline and the proposed w/o UBS were lower than those of the baseline (all data). These methods used a smaller size (by half) of the training data than the baseline (all data). This difference is due to class imbalances in the training data, as described in Section~\ref{sec:analysis} below. We used the two datasets with large class imbalances, where the number of samples decreases as the severity increases. The proposed method mitigated class imbalances in the training data and improved the performance by automatically selecting samples from minority classes.

In the results for ``Neighboring'' (difficult case), the proposed method outperformed all other methods in terms of mean accuracy for neighbor pairs. The performance comparison for each pair showed that the proposed method performed better than the other methods on ``Mayo~1--2'' and ``Mayo~2--3''. In particular, the proposed method improved the accuracy of ``Mayo~2--3'' by more than $10\%$ compared to the other methods on the private dataset. The more severe the disease, the greater the need for treatment, so evaluating treatment effects on severe diseases is critical. The proposed method performed better on severe image pairs and is thus considered superior to the other methods in terms of clinical application. However, the results of ``Mayo~0--1'' showed that the accuracy of the proposed method was lower than that of the other methods. This is because the proposed method mitigates the class imbalance in the training data and thus has fewer images labeled ``Mayo 0--1'' in the training data.

Figure~\ref{fig4} shows the change in the accuracy of estimating relative labels at each iteration (each labeling ratio) for the ``Overall'' and ``Neighboring'' test sets. The horizontal axis is the labeling ratio, the vertical axis is the mean accuracy for five-fold cross-validation, and the black dotted line is the baseline (all data) result with a labeling ratio of $100\%$. As we can see, the accuracy of the proposed method (red) improved as the amount of training data increased, and the improvement was more pronounced than that of the other methods. The accuracy of the proposed method was better than that of the baseline (all data) when the labeling ratio was $40\%$ in the ``Neighboring'' test set. In contrast, the baseline (blue), the proposed w/o UBS (green), and Core-set (orange) showed only marginal improvement in accuracy as the amount of training data increased, with limited gains from the initial training. As shown in Table~\ref{tb:accracy}, the accuracies of the compared methods were lower than that of the proposed method, especially for Mayo~2--3, which is a pair of minority classes. The lack of significant accuracy increase over iterations in the compared methods can be attributed to the effect of class imbalance in the pair creation. These results indicate that active learning with uncertainty-based sampling effectively selects highly effective pairs for learning. 

Figure~\ref{fig5} shows box plots of the estimated rank scores at each Mayo score from the baseline and the proposed method. The horizontal axis shows each Mayo score, and the vertical axis shows the mean estimated rank scores. The left figure, ``Initial'', shows the method where no iterations were performed, and the network was trained using only the initial training data (labeling ratio of $20\%$). The baseline and the proposed method refer to the results obtained using the training data with a labeling ratio of $50\%$. In the box plots, we consider the rank scores to be reasonable when there is little overlap in the distribution of rank scores at each Mayo score, as the estimated difference in severity is clear. In the initial and the baseline results, the rank score distributions of Mayo~2 and~3 overlapped significantly, while in contrast, the proposed method reduced the overlap. These results indicate that the estimated rank score and the Mayo score are highly correlated in the proposed method.

\subsubsection{Relationship between uncertainty and class imbalance}\label{sec:analysis}

As discussed in Section~\ref{sec:eval}, the proposed method significantly improved the accuracy of relative label estimation by automatically sampling from minority classes. These results suggest a relationship between the model uncertainty using the Bayesian CNN and the minority classes in the dataset. Therefore, we examined the number of samples of each Mayo score and the uncertainty of the sampled images in the training data updated by iterations.

Figure~\ref{fig6} shows the class proportions of the Mayo score of the samples in the training data of the baseline and the proposed method when the labeling ratio is $20\%$ ($k=0$), $35\%$ ($k=3$), and $50\%$ ($k=K=6$). The horizontal axis shows the Mayo scores, and the vertical axis shows the average number of images in the five-fold cross-validation. We analyzed the difference between uncertainty-based and random sampling by the number of images in each Mayo score. The initial training data in each method (labeling ratio of $20\%$) had the same class imbalance as the entire dataset due to random sampling. In addition, at the labeling ratios of $35\%$ and $50\%$ in the baseline, the training data had class imbalances similar to the initial training data. In the baseline, this class imbalance in the training data affected the performance. Thus, the performance improvement was limited despite increasing the number of images in the training data. In contrast, the proposed method gradually mitigated the class imbalance by selecting more minority class images (Mayo~2 and~3) as the number of iterations increased. As a result, the accuracy of relative label estimation was significantly improved, even though the amount of training data was half that of the baseline (all data).

Figure~\ref{fig7} shows the model uncertainty distribution of the sample at each Mayo score on the training data for the baseline and the proposed method. These results were obtained using the training data with a labeling ratio of $50\%$. The horizontal axis shows the Mayo scores, and the vertical axis shows the uncertainty of the samples in the training data. In the baseline, the uncertainty in the minority classes (Mayo~2 and~3) is higher than in the majority classes (Mayo~0 and~1). In the proposed method, the uncertainty of the sample at each Mayo score is not much different, and the uncertainty for Mayo~2 and~3 is lower than the baseline uncertainty. These results indicate that the uncertainty is correlated with the class imbalance, and the smaller the sample size of the class, the higher the uncertainty. Therefore, the proposed method achieved a high performance thanks to using class-balanced training data obtained by uncertainty-based sampling as discussed in Section~\ref{sec:eval}. The proposed method is thus useful for ranking tasks in medical images because serious class imbalance problems often occur in medical image datasets.

\subsubsection{Qualitative evaluation}\label{sec:qualit}
\begin{figure*}[t]
\centering
\includegraphics[scale=0.6]{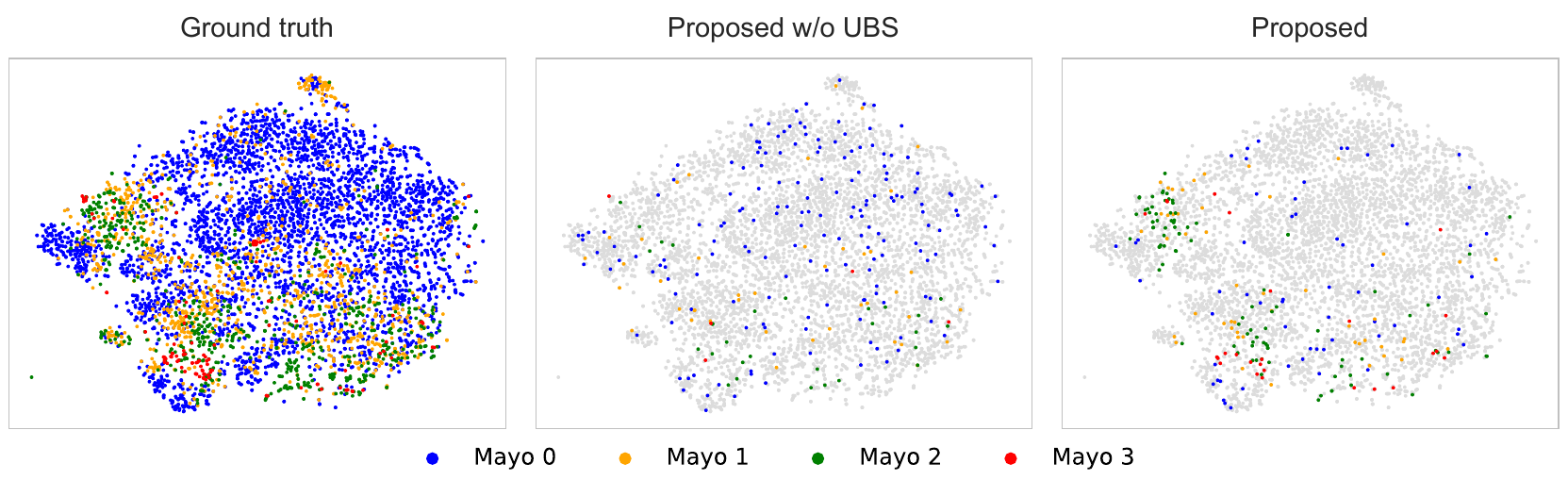}
\caption{$t$-SNE visualization of the feature distribution and the sampling results after initial training. In the middle and right panels, colored dots represent selected images.}
\label{fig8}
\end{figure*}

We examined the distribution of the samples selected by UBS in the feature space. Figure~\ref{fig8} shows the distribution of selected samples at the first iteration just after the initial training and Mayo scores in the feature space using $t$-SNE. In this visualization, we extracted 1664-dimensional features from images using DenseNet-169 trained on ImageNet and compressed these features into a two-dimensional map with $t$-SNE. The left panel shows the distribution of all training data and the corresponding Mayo scores. The middle and right panels show the differences in the distribution of selected samples using the proposed method, based on the presence or absence of UBS. The proposed method without UBS selected samples uniformly, predominantly choosing samples from Mayo~0, the majority class. In contrast, with UBS, the proposed method selected samples from minority classes more frequently, thereby reducing class imbalance.

\subsection{Multi-class classification}\label{sec:eval_class}
\subsubsection{Dataset}\label{sec:class_dataset}
To investigate the effectiveness of the proposed method in classification tasks, we conducted experiments to evaluate its performance for multi-class classification. In these experiments, we used the private dataset with large class imbalances, and the test data preserved the class imbalance. We also used five-fold cross-validation, and the dataset was divided into training ($60\%$), validation ($20\%$), and test ($20\%$) sets with patient-based sampling.

\subsubsection{Evaluation metrics}\label{sec:class_evaluation}
We used precision, recall, and F1-score as performance metrics for multi-class classification. We evaluated classification performance primarily on F1-scores because the dataset has a large class imbalance. The rank scores obtained from the trained CNN are not categorical scales and cannot be used in classification evaluations without modification. Therefore, we determined the class by quantizing the rank score to the nearest integer for converting the rank scores to the discrete severity classes (Mayo scores). For example, a rank score of 1.7 is classified as Mayo~2. 

\subsubsection{Compared methods}\label{sec:class_comparison}
We compared the proposed method with baseline, baseline (all data), and proposed w/o UBS in Section~\ref{sec:comparison}, and a conventional CNN-based method. In the conventional method, the CNN backbone used DenseNet-169 (the same as the proposed method) and was trained with categorical cross-entropy as the loss function. All 8,214 training samples with absolute labels (Mayo scores) were used to train the conventional method.

\subsubsection{Annotation efficiency evaluation}\label{sec:annotate_effici}
\begin{table}[t]
  \centering
  \caption{Comparison of the number of relative and absolute labels and the annotation time for training data after additional absolute annotations.}
  \label{tb:annotation time}
    \scalebox{1}{
    \begin{threeparttable}
    \begin{tabular}{lrrr}
     \hline
     \multirow{2}{*}{Method}&\multicolumn{2}{c}{Labels}&\multirow{2}{*}{Time (s)\tnote{*}}\\
     \cline{2-3}
     &Relative&Absolute&\\
     \hline \hline
     Conventional&0&8,214&164,280\\
     Baseline\tnote{**}&4,106&4,106&86,226\\
     Baseline (all data)&8,214&8,214&172,494\\
     Proposed w/o UBS\tnote{**}&4,106&3,378&71,666\\
     Proposed\tnote{**}&4,106&{\bf2,475}&{\bf53,606}\\
     \hline 
    \end{tabular}
    \begin{tablenotes}
    \item[*]Relative: 1 (s$/$pair), Absolute: 20 (s$/$image)
    \item[**]Relative labeling ratio of 50\% 
    \end{tablenotes}
    \end{threeparttable}
    }
\end{table}

\begin{table}[t]
  \centering
  \caption{Classification performance evaluation on test data.}
  \label{tb:Class_evalu}
    \scalebox{1}{
    \begin{tabular}{lccc}
     \hline
     Method&Precision&Recall&F1-score\\
     \hline \hline
     Conventional&0.626&0.642&0.629\\
     Baseline&0.661&0.623&0.627\\
     Baseline (all data)&0.632&{\bf0.655}&0.640\\
     Proposed w/o UBS&0.629&0.634&0.620\\ 
     Proposed&{\bf0.682}&0.641&{\bf0.649}\\ 
     \hline
    \end{tabular}
    }
\end{table}

Table~\ref{tb:annotation time} shows the relative and absolute labels and annotation costs of the training data for each method in the classification tasks. We used the training data with a relative labeling ratio of $50\%$ in the baseline, the proposed w/o UBS, and the proposed method. Annotation times were calculated as follows. The relative annotation takes one second per pair, and the absolute annotation takes 20 seconds per image~\citep{Kadota_Access_2022}. As shown in Table~\ref{tb:annotation time}, the annotation time of the proposed method was less than that of all other methods and was as low as about one-third of that of the conventional method. Interestingly, the proposed method had about 900 fewer absolute labels than the proposed w/o UBS. The proposed method preferentially selects minority class images with the highest learning effect to create pairs. We believe the number of absolute labels was reduced because the proposed method repeatedly selected the same minority class images during active learning. Note that the proposed method always generates non-duplicate pairs and performs relative annotation, even if active learning selects the same image repeatedly.

\subsubsection{Classification performance evaluation}\label{sec:class_perfor_evalua}
Table~\ref{tb:Class_evalu} shows the classification performance of each method. As we can see, the F1-score of the proposed method was higher than all other methods. In particular, it performed better than the baseline (all data) trained with all training images. We found that the proposed method can preferentially select the more effective images for learning in classification. In addition, the proposed method achieved a higher F1-score than the conventional method. These findings demonstrate the superiority of the proposed method in terms of significantly reducing the annotation cost for disease severity classification.

%--------------------------------------------------------
\section{Conclusion}
%--------------------------------------------------------
We proposed a deep Bayesian active learning-to-rank for efficient relative annotation by uncertainty-based sampling. We theoretically proved that MC dropout can be applied to estimate the model uncertainty of a pairwise LTR using a Bayesian Siamese neural network. The proposed method actively determines effective sample pairs for additional relative annotation by estimating the model uncertainty of the samples using Bayesian CNN. Experimental results showed that the proposed method achieves high performance with a small number of samples by selecting highly effective images for learning in ranking and classification tasks. We also found that it is robust to class imbalances because it selects a minority but significant samples. The proposed method has two drawbacks. First, it is computationally expensive because it requires multiple estimations of output values to obtain uncertainty. In future work, we plan to investigate an active learning-to-rank approach that allows for multiple estimations in a short time. Second, uncertainty may become inaccurate for datasets from domains that are different from the training dataset. We will explore domain generalization for an active learning-to-rank approach that calibrates uncertainty across multiple domains.

\section*{Declaration of Competing Interest}
None.

\section*{CRediT authorship contribution statement}
{\bf Takeaki Kadota:} Conceptualization, Formal analysis, Funding acquisition, Investigation, Methodology, Software, Validation, Visualization, Writing - original draft. {\bf Hedeaki Hayashi:} Conceptualization, Formal analysis, Funding acquisition, Methodology, Supervision, Writing - review \& editing. {\bf Ryoma Bise:} Funding acquisition, Methodology, Supervision, Writing - review \& editing. {\bf Kiyohito Tanaka:} Resources, Data curation. {\bf Seiichi Uchida:} Funding acquisition, Methodology, Project administration, Resources, Supervision, Writing - review \& editing.

\section*{Acknowledgments}
This work was supported by JSPS KAKENHI Grant Numbers JP20H04211, JP21H03511, and JP23K18509, AMED Grant Number JP20lk1010036h0002, and JST SPRING Grant Number JPMJSP2136.

%%Harvard
\bibliographystyle{model2-names.bst}\biboptions{authoryear}
\bibliography{main}

\end{document}